\documentclass[runningheads, envcountsame, a4paper]{llncs}
\usepackage{array}
\usepackage{float}
\usepackage{multirow}
\usepackage{amsmath}
\usepackage{amssymb}
\usepackage[pdftex]{graphicx}
\usepackage{epstopdf}
\usepackage[hyphens]{url}

\makeatletter

\providecommand{\tabularnewline}{\\}

\author{Phuoc Nguyen\inst{1} \and
Truyen Tran\inst{1} \and
Sunil Gupta\inst{1} \and
Santu Rana\inst{1} \and
Hieu-Chi Dam\inst{2} \and
Svetha Venkatesh\inst{1}}

\tocauthor{Phuoc~Nguyen, Truyen~Tran, Sunil~Gupta, Santu~Rana, Hieu-Chi~Dam, Svetha~Venkatesh}

\authorrunning{P. Nguyen et al.}
\institute{A2I2, Deakin University, Geelong, Australia \\
\email{\{phuoc.nguyen, truyen.tran, sunil.gupta, santu.rana, svetha.venkatesh\}@deakin.edu.au} \and
Japan Advanced Institute of Science and Technology, Japan \\
\email{dam@jaist.ac.jp}}

\makeatother

\begin{document}
\title{Variational Hyper-Encoding Networks}
\toctitle{Variational Hyper-Encoding Networks}

\maketitle
\setcounter{footnote}{0}
\begin{abstract}
We propose a framework called HyperVAE for encoding distributions
of distributions. When a target distribution is modeled by a VAE,
its neural network parameters are sampled from a distribution in the
model space modeled by a hyper-level VAE. We propose a variational
inference framework to implicitly encode the parameter distributions
into a low dimensional Gaussian distribution. Given a target distribution,
we predict the posterior distribution of the latent code, then use
a matrix-network decoder to generate a posterior distribution for
the parameters. HyperVAE can encode the target parameters in full
in contrast to common hyper-networks practices, which generate only
the scale and bias vectors to modify the target-network parameters.
Thus HyperVAE preserves information about the model for each task
in the latent space. We derive the training objective for HyperVAE
using the minimum description length (MDL) principle to reduce the
complexity of HyperVAE. We evaluate HyperVAE in density estimation
tasks, outlier detection and discovery of novel design classes, demonstrating
its efficacy.

\end{abstract}

\global\long\def\E{\mathbb{E}}%
\global\long\def\R{\mathbb{R}}%
\global\long\def\KL{D_{KL}}%
\global\long\def\BO{\text{BO}}%
\global\long\def\defeq{\stackrel{\text{def}}{=}}%
\global\long\def\Dcal{{\cal D}}%
\global\long\def\Ncal{\mathcal{N}}%
\global\long\def\Mcal{{\cal M}}%
\global\long\def\Xcal{{\cal X}}%
\global\long\def\Zcal{{\cal Z}}%
\global\long\def\Ucal{{\cal U}}%
\global\long\def\degree{^{\circ}\text{C}}%
\global\long\def\argmax{\operatornamewithlimits{argmax}}%
\global\long\def\argmin{\operatornamewithlimits{argmin}}%

\global\long\def\Ebb{\mathbb{E}}%
\global\long\def\Hbb{\mathbb{H}}%
\global\long\def\Ibb{\mathbb{I}}%
\global\long\def\Rbb{\mathbb{R}}%
\global\long\def\d{\text{d}}%

\global\long\def\Ncal{\mathcal{N}}%
\global\long\def\Mcal{{\cal M}}%
\global\long\def\Tcal{{\cal T}}%
\global\long\def\Lcal{{\cal L}}%
\global\long\def\Pcal{{\cal P}}%

\global\long\def\erf{\text{Erf}}%

\section{Introduction}

Humans can extract meta knowledge across tasks such that when presented
with an unseen task they can use this meta knowledge, adapt it to
the new context and quickly solve the new task. Recent advance in
meta-learning \cite{finn2017model,grant2018recasting}
shows that it is possible to learn a single model such that when presented
with a new task, it can quickly adapt to the new distribution and
accurately classify unseen test points. Since meta-learning algorithms
are designed for few-shot or one-shot learning where labeled data
exists, it faces challenges \emph{when there is none}\footnote{This is not the same as zero-shot learning where label description
is available.} to assist backpropagation when testing.

Hyper-networks \cite{ha2016hypernetworks} can generate the weights
for a target network given a set of embedding vectors of those weights.
Due to its generative advantage, it can be used to generate a distribution
of parameters for a target network \cite{ha2016hypernetworks,krueger2017bayesian}.
In practice, due to the high dimensional parameter space, it only
generates scaling factors and biases for the target network. This
poses a problem that the weight embedding vectors only encode partial
information about the target task, and thus are not guaranteed to
perform well on unseen tasks.

On the other hand, variational autoencoders (VAEs) \cite{kingma2013auto,rezende2014stochastic}
is a class of deep generative models that can model complex distributions.
A major attractive feature of VAEs is that we can draw from simple,
low-dimensional distributions (such as isotropic Gaussians), and the
model will generate high-dimensional data instantly without going
through expensive procedures like those in the classic MCMC. This
suggests VAEs can be highly useful for high dimensional design exploration
\cite{gomez2018automatic}. In this work, we lift this idea to one
more abstraction level, that is, using a \emph{hyper VAE to generate
VAE models}. While the VAEs work at the individual design level, the
\emph{hyper VAE} works at the class level. This permits far more flexibility
in exploration, because not only we can explore designs within a class,
we can explore multiple classes. The main insight here is that the
model parameters can also be treated as a design in a model design
space. Hence, we can generate the model parameters using another VAE
given some latent low-dimensional variable.

We propose HyperVAE, a novel class of VAEs, as a powerful deep generative
model to learn to generate the parameters of VAE networks for modeling
the distribution of different tasks. The versatility of the HyperVAE
to produce VAE models allows it to be applied for a variety of problems
where model flexibility is required, including density estimation,
outlier detection, and novelty seeking. For the latter, since HyperVAE
enforces a smooth transition in the model family, interpolating in
this space will enable us to extrapolate to models of new tasks which
are \emph{close} to trained tasks. Thus as global search techniques
can guide the generation of latent spaces of VAEs, search enables
HyperVAE to produce novel classes of discovery. We use Bayesian Optimization
(BO) \cite{shahriari2016taking},  to search in the low dimensional
encoding space of VAE. Once a low dimensional design is suggested,
we can decode it to the corresponding high dimensional design.

We demonstrate the ability of HyperVAE on
three tasks: density estimation, robust outlier detection and discovery
of unseen design classes. Our main contributions and results are:
(i) Development of a hyper-encoding framework, guided through MDL;
(ii) Construction of a versatile HyperVAE model that can tackle density
estimation tasks and outlier detection; and (iii) Demonstration of
novel designs produced from our model coupled with BO.

\section{Variational Autoencoder (VAE)}

Let $x$ denote an
$\Xcal$-value random variable associated with a $\Zcal$-value random
variable $z$ through a joint distribution $p(x,z)$. We consider
a parametric family $\Pcal$ of generative models $p(x,z;\theta)$
factorized as a conditional $p(x|z;\theta)$ and a simple prior $p(z)$,
usually chosen as $\Ncal(0,I)$. Maximum likelihood estimate (MLE)
of $\theta\in\Theta$, where $\Theta$ is the parameter space, over
the marginal $\log p(x;\theta)=\log\int p(x,z;\theta)dz$ is intractable,
thus requiring alternatives such as expectation-maximization and variational
inference. VAE is an amortized variational inference that
jointly learns the generative model $p(x|z;\theta)$ and the variational
inference model $q(z|x;\theta)$\footnote{We use $\theta=(\theta_{p},\theta_{q})$ to denote the set of parameters
for $p$ and $q$.}. Its ELBO objective,
\begin{equation}
\Lcal(x,p,q;\theta)=\E_{q(z|x;\theta)}\log p(x|z;\theta)-\KL\left(q(z|x;\theta)\|p(z)\right)\label{eq:elbo-1}
\end{equation}
\noindent lower-bounds the marginal log-likelihood $\log p(x;\theta)$.
In practice, Monte Carlo estimate of the ELBO's gradient is used
to update $\theta$. The form of $q$ and $p$ in Eq.~\ref{eq:elbo-1}
makes an encoder and a decoder, hence the name auto-encoder \cite{kingma2013auto}.

\section{Variational Hyper-Encoding Networks}

We assume a setting where there is a sequence of datasets (or tasks)
and model parameters $\{(D_{t},\theta_{t})\}_{t}$ a sender wish to
transmit to a receiver using a minimal combined code length.

\subsection{Hyper-auto-encoding problem}

Given a set of $T$ distributions $\{D_{t}\}_{t=1}^{T}$ called tasks,
each containing samples $x\sim p_{D_{t}}(x)$, our problem is first
fitting each parametric model $p(x;\theta_{t})$, parameterized by
$\theta_{t}\in\Theta$, to each $D_{t}$:
\begin{equation}
\hat{\theta}_{t}=\argmax_{\theta\in\Theta}p(D_{t};\theta)\label{eq:mle}
\end{equation}
\noindent then fitting a parametric model $p(\theta;\gamma)$, parameterized
by $\gamma\in\Gamma$ to the set $\{\hat{\theta}_{t}\}_{t=1}^{T}$.
However, there are major drawbacks to this approach. First, the number
of tasks may be insufficient to fit a large enough number of $\theta_{t}$
for fitting $p(\theta;\gamma)$. Second, although we may resample
$D_{t}$ and refit $\theta_{t}$ to create more samples, it is computationally
expensive. A more practical approach is to jointly learn the distribution
of $\theta$ and $D$.

\subsection{Hyper-encoding problem }

Our problem is to learn the joint distribution $p(\theta,D;\gamma)$
for some parameters $\gamma$\footnote{We assume a Dirac delta distribution for $\gamma$, i.e. a point estimate,
in this study.}.

\paragraph{HyperVAE}

\begin{figure}
\begin{centering}
\begingroup\tabcolsep=0pt\def\arraystretch{0}%
\begin{tabular}{ccc}
\includegraphics[viewport=0bp 0bp 200bp 150bp,clip,height=2.9cm]{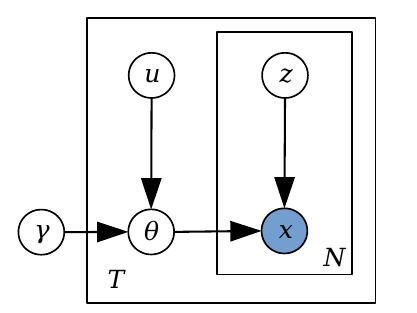} & \includegraphics[viewport=0bp 0bp 210bp 150bp,clip,height=2.9cm]{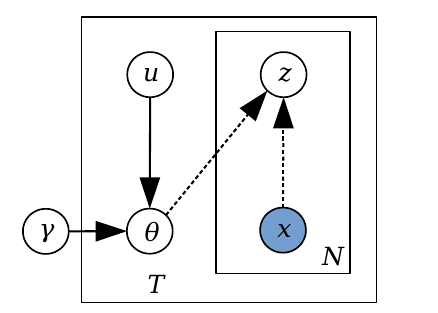} & \includegraphics[viewport=0bp 0bp 210bp 150bp,clip,height=2.9cm]{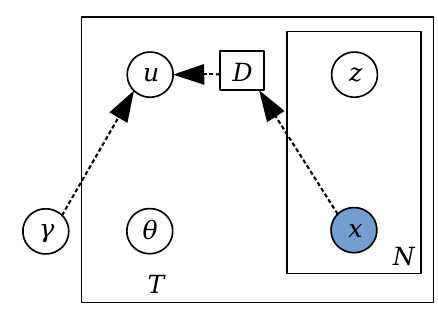}\tabularnewline
(a) Generation & (b) Inference of $z$ & (c) Inference of $u$\tabularnewline
\end{tabular}\endgroup
\par\end{centering}
\caption{HyperVAE networks, $D=\{x_{n}\}_{n=1}^{N}$.
\label{fig:pgm}}
\end{figure}

We propose a framework for this problem called \emph{HyperVAE} as
depicted in Fig.~\ref{fig:pgm}. The main insight here is that
the \emph{VAE model parameters} $\theta\in\Theta$ can also be treated
as a normal input in the parameter space $\Theta$. Hence, we can
generate the model parameters $\theta$ using \emph{another VAE at
the hyper level} whose generative process is $p_{\gamma}(\theta|u)$
for some low-dimensional latent variable $u\sim p(u)\equiv\Ncal(0,I)$,
the prior distribution defined over the latent manifold $\Ucal$ of
$\Theta$. The joint distribution $p(\theta,D;\gamma)$ can be expressed
as the marginal over the latent representation $u$:
\begin{align}
p(\theta,D) & =\int p(\theta,D|u)p(u)\d u=\int p(D|\theta)p(\theta|u)p(u)\d u\label{eq:hyper-encoding-problem}
\end{align}

\paragraph{Generation}

\noindent of a random data point $x$ is as follows, c.f. , Fig.~\ref{fig:pgm}~(a):
\begin{align*}
u_{t} & \sim N(0,I)\\
\theta_{t} & \sim p_{\gamma}(\theta\mid u_{t})\\
z & \sim N(0,I)\\
x & \sim p_{\theta_{t}}(x\mid z)
\end{align*}
\paragraph{Inference of $z$}
given $x$ and $\theta$, Fig.~\ref{fig:pgm}~(b), is approximated
by a Gaussian distribution, $q(z|x,\theta)=\Ncal(z|\mu_{\theta}(x),\sigma_{\theta}^{2}(x))$,
where $\mu_{\theta}$ and $\sigma_{\theta}^{2}$ are neural networks
generating the mean and variance parameter vectors.
\paragraph{Inference of $u$}
is shown in Fig.~\ref{fig:pgm}~(c). We also assume a Gaussian posterior
distribution $q(u|D,\theta)=\Ncal(u|\mu(d_{t}),\sigma^{2}(d_{t}))$
parameterized by neural networks $\mu(.)$ and $\sigma^{2}(.)$. Since
$\theta$ can be trained on $D$ thus depending on D, we can approximate
this dependency implicitly using the inference network itself, thus
$q(u|D,\theta)\approx q(u|D)$. The next problem is that since $D_{t}$
is a set, $q(u|D_{t})$ is a function of set, which is an interesting
problem on its own. Here we use a simple method to summarize $D_{t}$
into a vector,
\begin{equation}
d_{t}=s(D_{t})\label{eq:set-task}
\end{equation}
and this turns $q(u|D_{t})$ into $q(u|d_{t})$. For example, $s(\cdot)$
can be a mean function, a random draw from the set, or a description
of the set. In this study, we simply choose a random draw $x$ from
the set $D_{t}$.

\subsection{Minimum Description Length}

It is well-known that variational inference is equivalent to the Minimum
Description Length (MDL) principle \cite{hinton1993keeping}. In this
section, we use MDL to compute the total code length of the model
and data misfits. From the total code length, we show that a shorter
code length and a simpler model can be achieved by redesigning the
distribution of the model space. We used a Dirac delta distribution
centered at $\mu(u)$ for $\theta$ given each latent code $u$, $p(\theta|u)=\delta_{\mu(u)}(\theta)$
parameterized by a neural network $\mu(u)$ for each latent code $u$.
This results in an implicit distribution for $\theta$ represented
by the compound distribution $p(\theta)=\int\delta_{\mu(u)}(\theta)p(u)\d u$.

Under the crude 2-part code MDL, the expected code length for transmitting
a dataset $D$ and the model parameters $\theta$ in the encoding
problem in Eq.~\ref{eq:mle} is $L(D)=L(D|\theta)+L(\theta)$, where
$L(D|\theta)=-\log p(D;\theta)$\footnote{We abused the notation and use $p$ to denote both a density and a
probability mass function. Bits-back coding is applicable to continuous
distributions \cite{hinton1993keeping}.} is the code length of the data given the model $\theta$, and $L(\theta)=-\log p(\theta)$
is the code length of the model itself. Under the HyperVAE the code
length of $D$ is:
\begin{align}
L(D) & =L(D|\theta)+L(\theta|u)+L(u)\label{eq:naive-hyper}
\end{align}

If we choose a Dirac delta distribution for $\theta$, $p(\theta|u)=\delta_{\mu(u)}(\theta)$
then $\theta$ is deterministic from $u$ and we can eliminate the
code length $L(\theta|u)$, thus making the total code length shorter:
\begin{align}
L(D) & =L(D|\theta(u))+L(u)\label{eq:dirac-hyper}
\end{align}
Additionally, bits-back coding can recover the additional information
in the entropy of the variational posterior distribution $q(u|D)$,
thus this information should be subtracted from the total code length
\cite{hinton1993keeping,townsend2019practical}. The total expected
code length is then:
\begin{align}
L(D) & =\E\left[L\left(D|\theta(u)\right)-\log p(u)+\log q(u|D)\right]\nonumber \\
 & =\E\left[L\left(D|\theta(u)\right)\right]+\KL\left(q(u|D)\|p(u)\right)\label{eq:mdl-data}
\end{align}
where the expectation is taken over the posterior distribution $q(u|D)$.
The description length of a dataset $D=\{x_{i}\}_{i=1}^{|D|}$ is
the summation of the description length of every data point:
\begin{align}
L\left(D|\theta(u)\right) & =\sum_{i=1}^{|D|}L\left(x_{i}|\theta(u)\right)\nonumber \\
 & =\sum_{i=1}^{|D|}\left(\E_{q_{\theta}(z|x_{i})}L\left(x_{i}|z,\theta\right)+\KL\left(q_{\theta}(z|x_{i})\|p(z)\right)\right)\label{eq:elbo-D}
\end{align}
where we ignored the dependence of $\theta$ on $u$ to avoid clutter.
We train the HyperVAE parameters by minimizing the description length
in Eq.~\ref{eq:mdl-data}. In our experiment, we scale down this
objective by multiplying it by $1/|D|$ to have a similar scale as
a normal VAE's objective. The training objective for HyperVAE is then:
\begin{align}
L(D) & =\frac{1}{|D|}\left[L\left(D|\theta(u)\right)+\KL\left(q(u|D)\|p(u)\right)\right]\label{eq:training-obj}
\end{align}

\paragraph{Mini-batches as tasks}

In practice, the number of tasks is too small to adequately train
the hyper-parameters $\gamma$. Here we simulate tasks using data
mini-batches in the typical stochastic gradient learning. That is,
each mini-batch is treated as a task. To qualify as a task, each mini-batch
needs to come from the same class. For example, for handwritten digits,
the class is the digit label.

\subsection{Compact hyper-decoder architecture}

Since neural networks weights are matrices that are highly structured
and often overparameterized, we found that a more efficient method
is to use a matrix generation network \cite{do2017matrix} for generating
the weights. More concretely, a matrix hyper-layer receives an input
matrix $H$ and computes a weight matrix $W$ as $W=\sigma(UHV+B)$,
where $U,V,B$ are parameters. As an example, if $H$ is a 1D matrix
of size $400\times1$ and a target weight $W$ of size $400\times400$,
a matrix-layer will require $176$ thousand parameters, a 3 order
of magnitude reduction from $64.16$ million parameters of the standard
fully-connected hyper-layer. This compactness allows for complex decoder
architecture for generating the target network, unlike hyper-networks
methods which rely on a linear layer of an embedding vector for each
target-network layer.

\subsection{Applications}

We can use the HyperVAE framework for density estimation, outlier
detection and novelty discovery. In the following, we use \emph{HyperVAE
}to denote the whole VAE-of-VAEs framework, \emph{hyper VAE} for the
hyper level VAE, and \emph{main VAE} for the VAE of each target task.
\paragraph{Density estimation}
After training, HyperVAE can be used to estimate the density of a
new dataset/task. Let $D_{t}$ is the new task data. We first infer
the posterior distribution $q(u|D_{t})\approx q(u|d_{t})=\Ncal(u|\mu(d_{t}),\sigma^{2}(d_{t}))$,
where $d_{t}$ is a summary of $D_{t}$, Eq.~\ref{eq:set-task},
which we choose as random in this study. Next we select the mean of
this posterior distribution and decode it into $\theta$ using $p_{\gamma}(\theta|u)$.
We use this $\theta$ to create the main VAE for $D_{t}$ then use
importance sampling to estimate the density of $x\in D_{t}$ as follows:
\[
p(x)=\E_{q(z|x,\theta)}\frac{p(x|z)p(z)}{q(z|x,\theta)}\approx\frac{1}{N}\sum_{i=1}^{N}\frac{p(x|z_{i})p(z_{i})}{q(z_{i}|x,\theta)}
\]
where $N$ is a chosen number of importance samples, $\{z_{i}\}_{i=1}^{N}$
are samples from the proposal distribution $q(z|x,\theta)$ to reduce
the variance of the density estimate, and $p(z_{i})/q(z_{i}|x,\theta)$
is the multiplicative adjustment to compensate for sampling from $q(z|x,\theta)$
instead of $p(z)$.

\paragraph{Outlier detection}

Similar to the density estimation application above, we first encode
a test vector $x_{t}$ into a latent distribution $q(u|x_{t})$ then
decode its mean vector into $\theta_{t}$ to create a VAE model. We
then use the description length of $x_{t}$, c.f. Eq.~\ref{eq:elbo-D},
under this VAE as the outlier score. Our assumption is that outliers
are unseen to the trained model, thus incompressible under this model
and should have longer description lengths.

\paragraph{Novelty discovery}

HyperVAE provides an extra dimension for exploring the model space
$\Theta$ in addition to exploring the design space $\Xcal$. Once
trained, the network can guide exploration of new VAE models for new
tasks with certain similarity to the trained tasks.

Given no prior information, we can freely draw models $\theta(u)$
from $u\sim p(u)$ and designs $x\sim p_{\theta}(x|z)$ with $z\sim p_{\theta(u)}(z)$
and search for the desired $x^{*}$ satisfying some property $F(x^{*})$.
An intuitive approach is to employ a global search technique such
as Bayesian Optimization (BO) in both the model latent space of $u$
and in the data latent space of $z$. However searching for both $u\in\Ucal$
and $z\in\Zcal$ is expensive due to the combined number of dimensions
can be very high. Furthermore, reducing the latent dimension would
affect the capacity of VAE. To overcome this major challenge, we use
BO for optimizing the $z$ space and replace the search in $u$ space
by an iterative search heuristic. The workflow starts with an initial
exemplar $x_{0}^{*}$ which can be completely uninformative (e.g.,
an empty image for digits or a random design), or properly guided
(e.g., from the best choice thus far in the database, or from what
is found by VAE+BO itself). The search process for the optimal design
at step $t=1,2,...,T$ is as follows:
\begin{align}
u_{t} & \sim q(u\mid d_{t-1});\quad\quad\theta_{t}=g_{\gamma}(u_{t});\nonumber \\
z^{*} & \leftarrow\BO(g_{\theta_{t}}(z));\quad\quad x_{t}^{*}\leftarrow g(z^{*}).\label{eq:bo-step}
\end{align}
where $d_{t-1}\leftarrow x_{t-1}^{*}$.

The optimization step in the $z$ space maximizes a function $\text{\ensuremath{\max_{x}}}F(x)=\max_{z}F\circ g_{\theta_{t}}(z)$
for a fixed generator $\theta_{t}$. Let $z_{t}^{*}$ and thus $x_{t}^{*}=g_{\theta_{t-1}}(z_{t-1}^{*})$
be the solution found at step $t$. The generator parameter in the
subsequent step is set as $\theta_{t}\leftarrow\theta\left(\mu(x_{t}^{*})\right)$
where $\mu$ is the posterior mean. Thus the HyperVAE step transforms
the objective function with respect to $z$ by shifting $\theta$.
% The effect of the transformation is similar to that of landscape deformation
% in the global optimization \cite{hansmann2002global}, to help the
% search to escape from local maxima by deforming the shape of the minima.

\section{Experiments}

We evaluate HyperVAE on three tasks: density estimation, robust outlier
detection, and novel discovery.

\subsection{Data sets}

We use four datasets: MNIST handwritten digits, Omniglot handwritten
characters, Fashion MNIST, and Aluminium Alloys datasets. The MNIST
contains 60,000 training and 10,000 test examples of 10 classes ranging
from 0 to 9. The Omniglot contains 24,345 training and 8,070 test
examples. The Fashion MNIST dataset contains the same number of training
and test examples as well as the number of classes. In these three
datasets, the images are statically binarized to have pixel values
in $\{0,1\}$.

The Alloys dataset (https://tinyurl.com/tmah538), previously studied
in \cite{nguyen2019incomplete}, consists of 15,000 aluminium alloys.
Aluminium alloy is a combination of about $85\%$ aluminium and other
elements.

Phase diagram contains important characteristics of alloys, representing
variations between the states of compounds at different temperatures
and pressures. They also contain thermodynamic properties of the phases.
In this experiment, a phase diagram is coded as a 2D matrix, in which
each cell is the prevalence of a phase at a particular temperature.

\subsection{Model settings}

We use a similar architecture for the encoder and decoder of all VAE
in all datasets. The encoder has 2 convolution layers with 32 and
64 filters of size $3\times3$, stride 2, followed by one dense layer
with $100$ hidden units, then two parallel dense layers to output
the mean and log variance of $q(z|x;\theta)$. The decoder architecture
exactly reverses that of the encoder to map from $z$ to $x$, with
transposed convolution layers in place of convolution layers, and
outputs the Bernoulli mean of $p(x|z;\theta)$. For the alloys dataset,
the convolution layers are replaced by matrix layers with size $200\times200$,
as in \cite{do2018learning}. We also use a similar architecture for
HyperVAE in all datasets. The encoder uses the same architecture as
the VAE's encoder. The decoder use a dense layer with $100$ hidden
units, followed by $L$ parallel matrix layers generating the weights,
biases, and filters of the main VAE network, resembling the parameter
$\theta$. The input to the matrix layer is reshaped into size $20\times20$.
All layers except the last layer use RELU activation. The $z$-dimension
and $u$-dimension is 10 for all datasets. We used Adam optimizer
with parameters $\beta_{1}=0.9$, $\beta_{2}=0.999$, learning rate
$\eta=0.0003$, minibatches of size $100$, and ran for $10000$ iterations
or when the models converge.

\subsection{Model behavior}

We study whether the HyperVAE learns a meaningful latent representation
and data distribution for the MNIST and Omniglot datasets. We use
negative log-likelihood (NLL) and $\KL(q(z|x)\|p(z))$ as measures.
NLL is calculated using importance sampling with 1024 samples.
\begin{table}
\caption{Negative log-likelihood (-LL), and $\protect\KL(q(z|x)\|p(z))$ (KL).
Smaller values are better. \label{tab:Model-behavior}}

\centering{}%
\begin{tabular}{|>{\raggedright}p{1.5cm}|c|c|c|c|}
\hline
 &  & VAE & MetaVAE & HyperVAE\tabularnewline
\hline
\hline
\multirow{2}{1.5cm}{MNIST} & -LL & 99.4 & 93.0 & \textbf{88.2}\tabularnewline
\cline{2-5} \cline{3-5} \cline{4-5} \cline{5-5}
 & KL & 18.8 & 15.5 & 18.5\tabularnewline
\hline
\multirow{2}{1.5cm}{Omniglot} & -LL & 111.4 & 128.1 & \textbf{105.5}\tabularnewline
\cline{2-5} \cline{3-5} \cline{4-5} \cline{5-5}
 & KL & 17.1 & \textbf{13.2} & 18.1\tabularnewline
\hline
\multirow{2}{1.5cm}{Fashion MNIST} & -LL & 237.7 & 232.7 & \textbf{231.8}\tabularnewline
\cline{2-5} \cline{3-5} \cline{4-5} \cline{5-5}
 & KL & 14.5 & \textbf{13.7} & 13.9\tabularnewline
\hline
\end{tabular}
\end{table}

Table~\ref{tab:Model-behavior} compares the performance of VAE,
MetaVAE, and HyperVAE. As shown, HyperVAE has better NLL on the three
datasets. MetaVAE has smallest KL, which is due to it has a separate
and fixed generator for each task. HyperVAE has slightly smaller KL
on the MNIST and Fashion MNIST dataset than VAE. Note that better
log-likelihoods can be achieved by increasing the number of latent
dimensions, e.g. $\text{dim}(z)=50$, instead of $\text{dim}(z)=10$
in this experiment.

\begin{table}
\caption{Number of parameters (rounded to thousands). \label{tab:=000023param}}

\centering{}%
\begin{tabular}{|c|c|c|c|}
\hline
 & VAE & MetaVAE & HyperVAE\tabularnewline
\hline
\hline
Inference & 445 & $445$ & 445\tabularnewline
\hline
Generative & 445 & $445\times\#\text{task}$ & 445\tabularnewline
\hline
Total & 890 & $445+445\times\#\text{task}$ & 890\tabularnewline
\hline
\end{tabular}
\end{table}

Table~\ref{tab:=000023param} compares the number of parameters between
networks. Note that while MetaVAE shares the same inference network
for all tasks, it needs a separate generative network for each task.
For HyperVAE, the trainable parameters are from the hyper level VAE,
whereas the main VAE network of each task obtains its parameters by
sampling from the HyperVAE network. Therefore, for the comparison
we only count the number of trainable parameters, which is what eventually
saved to disk. The real parameters for the target networks will be
generated on-the-fly given a target task. Thus, it will take extra
generation time for each task, c.f. Table.~\ref{tab:Complexity}.
\begin{table}[H]
\caption{Time measured in milliseconds for a batch of 100 inputs.\label{tab:Complexity}}

\centering{}%
\begin{tabular}{|c|c|c|c|}
\hline
 & Generation & Inference & Total time\tabularnewline
\hline
\hline
VAE & 0.12 & 0.12 & 0.24\tabularnewline
\hline
MetaVAE & 0.12 & 0.12 & 0.24\tabularnewline
\hline
\multirow{2}{*}{HyperVAE} & 0.12 ($x$) & 0.12 ($z$) & \multirow{2}{*}{1.11}\tabularnewline
\cline{2-3} \cline{3-3}
 & 0.75 ($\theta$) & 0.12 ($u$) & \tabularnewline
\hline
\end{tabular}
\end{table}
\paragraph{Overfitting}

VAE is trained on the combined dataset therefore it is less affected
by overfitting due to high variance in the data. Whereas MetaVAE is
more susceptible to overfitting when the number of examples in the
target task is small, which is the case for Omniglot dataset, c.f.
Table~\ref{tab:Model-behavior}. While the training of MetaVAE's
encoder is amortized across all datasets, the training of its decoder
is task-specific. As a result, when a (new) task has a small number
of examples, the low variance data causes overfitting to this task's
decoder. Therefore MetaVAE is not suitable for transfer learning to
new/unseen tasks. HyperVAE can avoid overfitting by taking a Bayesian
approach.

\paragraph{HyperVAE complexity}

The algorithmic complexity of HyperVAE is about double that of VAE,
since it is a VAE of VAE, plus the extra generation time of the parameters.
Specifically, it runs the VAE at the hyper level to sample a weight
parameter $\theta$, then it runs the VAE to reconstruct a set of
inputs given this parameter $\theta$. Due to the difference in matrix
sizes of different layers in the target network, we generate each
weight matrix and bias vector at a time, resulting in $O(D)$ time\footnote{We assumed a matrix multiplication takes $O(1)$ time in GPU.}
with $D$ being the depth of the target network. Therefore, the time
complexity of the hyper generation network is $O(L_{\text{hyper}}+L_{\text{VAE}}+D)$,
where $L_{\text{hyper}}$ and $L_{\text{VAE}}$ are the number of
layers of the hyper and the primary generation networks respectively,
and we assumed the average hidden size of the layers is a constant.
However, more efficient methods is also possible. For example, inspired
from \cite{ha2016hypernetworks}, we can reshape matrices into batches
of blocks of the same size, then stacking along the batch dimension
in to a large 3D tensor. Then, we can use a matrix network to generate
this tensor in $O(1)$ time\footnote{Batched matrix multiplication can be paralleled in GPU.}
whence the time complex will be $O(L_{\text{hyper}}+L_{\text{VAE}})$.
We leave this implementation for future investigation. Table~\ref{tab:Complexity}
shows the wall-clock time comparison between methods on a Tesla P100
GPU.
\begin{table}
\caption{Outlier detection on MNIST. AUC: Area Under ROC Curve, FPR: False
Positive Rate, FNR: False Negative Rate, KL: KL divergence, -EL: mean
negative loglikelihood and KL. \label{tab:Outlier-detection}}

\centering{}%
\begin{tabular}{|>{\raggedright}m{1.4cm}|c|c|c|c|c|}
\hline
\multicolumn{1}{|>{\raggedright}m{1.4cm}}{\multirow{2}{1.4cm}{}} & \multirow{2}{*}{} & \multirow{2}{*}{VAE} & \multicolumn{2}{c|}{MetaVAE} & \multirow{2}{*}{HyperVAE}\tabularnewline
 &  &  & \multicolumn{1}{c}{KL} & -EL & \tabularnewline
\hline
\multirow{3}{1.4cm}{MNIST} & AUC & 93.0 & 54.7 & 52.2 & \textbf{95.3}\tabularnewline
\cline{2-6} \cline{3-6} \cline{4-6} \cline{5-6} \cline{6-6}
 & FPR & 16.3 & 47.5 & 49.4 & \textbf{15.6}\tabularnewline
\cline{2-6} \cline{3-6} \cline{4-6} \cline{5-6} \cline{6-6}
 & FNR & 15.5 & 45.0 & 50.5 & \textbf{8.0}\tabularnewline
\hline
\multirow{3}{1.4cm}{Omniglot} & AUC & 98.3 & 87.3 & 97.5 & \textbf{98.7}\tabularnewline
\cline{2-6} \cline{3-6} \cline{4-6} \cline{5-6} \cline{6-6}
 & FPR & 5.5 & 18.8 & 7.2 & \textbf{4.9}\tabularnewline
\cline{2-6} \cline{3-6} \cline{4-6} \cline{5-6} \cline{6-6}
 & FNR & 6.4 & 20.9 & 9.0 & \textbf{5.9}\tabularnewline
\hline
\multirow{3}{1.4cm}{Fashion MNIST} & AUC & 74.6 & 58.2 & 56.8 & \textbf{76.8}\tabularnewline
\cline{2-6} \cline{3-6} \cline{4-6} \cline{5-6} \cline{6-6}
 & FPR & \textbf{33.5} & 44.1 & 45.8 & 33.6\tabularnewline
\cline{2-6} \cline{3-6} \cline{4-6} \cline{5-6} \cline{6-6}
 & FNR & 32.0 & 43.5 & 44.5 & \textbf{28.7}\tabularnewline
\hline
\end{tabular}
\end{table}

\subsection{Robust outlier detection}

Next, we study HyperVAE model for outlier detection tasks. We use
three datasets: MNIST, Omniglot, and Fashion MNIST to create three
outlier detection experiments. For each experiment, we select one
dataset as the normal class and 20\% random samples from another dataset
as outliers. All methods are trained on only normal data. VAE and
HyperVAE use the negative log-likelihood and KL for calculating the
outlier score, which is equivalent to the negative ELBO. MetaVAE does
not have a generative network for new data. Therefore we use two scoring
methods: (1) KL divergence only, and (2) mean negative log-likelihood
and KL, using all trained generative networks. For training MetaVAE
and HyperVAE, we define the task as before, i.e. the data in each
task have a similar class label.

Table~\ref{tab:Outlier-detection} compares the performance of all
methods on the three datasets. Overall, HyperVAE has better AUCs compared
to VAE and MetaVAE. The MetaVAE has the lowest AUC. This could be
due to the use of a discrete set of generative networks for each task,
making it unable to handle new, unlabeled data.

While the false positive rates of VAE and HyperVAE models are similar,
the false negative rates for HyperVAE are lower than that of VAE.
This is because HyperVAE was trained across tasks, thus it has a better
support between tasks.
\begin{figure}
\begin{centering}
\begin{tabular}[b]{>{\centering}p{1.2cm}|c|c}
\hline
Novel digit & VAE & Iterations of HyperVAE\tabularnewline
\hline
1 & \includegraphics[height=0.5cm]{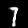} & \includegraphics[height=0.5cm]{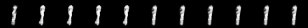}\tabularnewline
2 & \includegraphics[height=0.5cm]{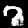} & \includegraphics[height=0.5cm]{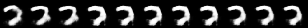}\tabularnewline
3 & \includegraphics[height=0.5cm]{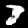} & \includegraphics[height=0.5cm]{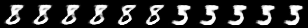}\tabularnewline
4 & \includegraphics[height=0.5cm]{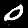} & \includegraphics[height=0.5cm]{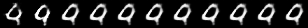}\tabularnewline
5 & \includegraphics[height=0.5cm]{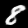} & \includegraphics[height=0.5cm]{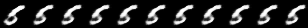}\tabularnewline
6 & \includegraphics[height=0.5cm]{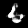} & \includegraphics[height=0.5cm]{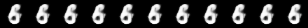}\tabularnewline
7 & \includegraphics[height=0.5cm]{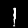} & \includegraphics[height=0.5cm]{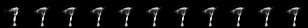}\tabularnewline
8 & \includegraphics[height=0.5cm]{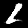} & \includegraphics[height=0.5cm]{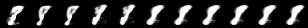}\tabularnewline
9 & \includegraphics[height=0.5cm]{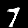} & \includegraphics[height=0.5cm]{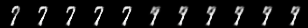}\tabularnewline
0 & \includegraphics[height=0.5cm]{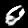} & \includegraphics[height=0.5cm]{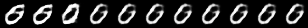}\tabularnewline
\hline
\end{tabular}
\par\end{centering}
\caption{Best digits found at iterative steps in searching for a new class
of digits, corresponding to the performance curves in Fig.~\ref{fig:Performance-curves-iterative-explore}.
\label{fig:Best-digits-found}}
\end{figure}

\begin{figure}
\begingroup\tabcolsep=0pt\def\arraystretch{0}%
\begin{tabular}{cccc}
\includegraphics[width=0.25\textwidth]{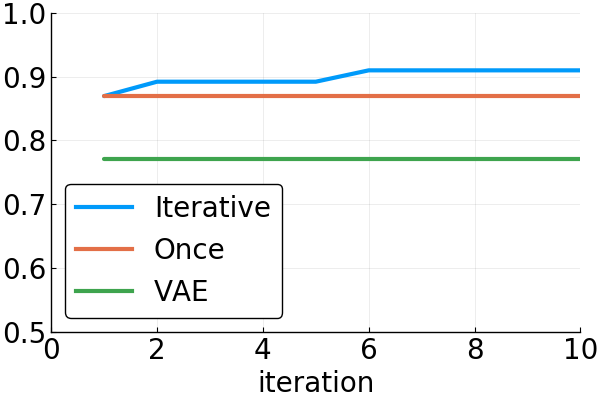} & \includegraphics[width=0.25\textwidth]{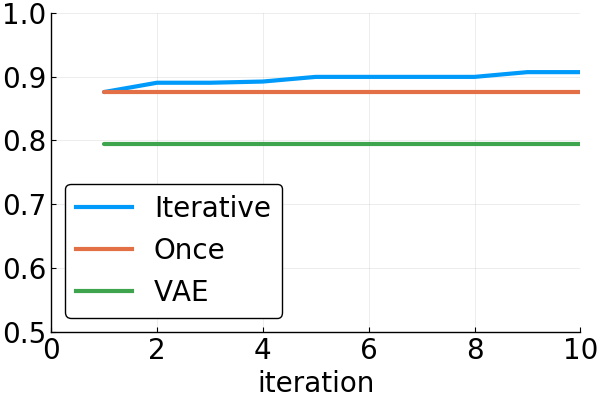} & \includegraphics[width=0.25\textwidth]{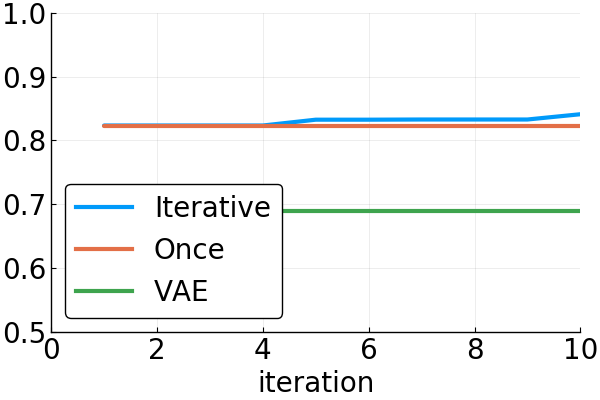} & \includegraphics[width=0.25\textwidth]{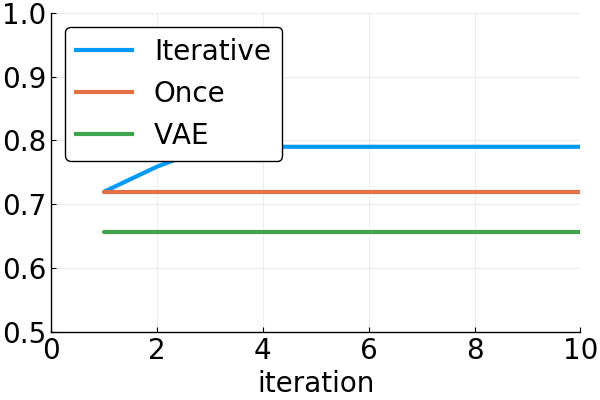}\tabularnewline
 &  & \medskip{}
 & \tabularnewline
\includegraphics[width=0.25\textwidth]{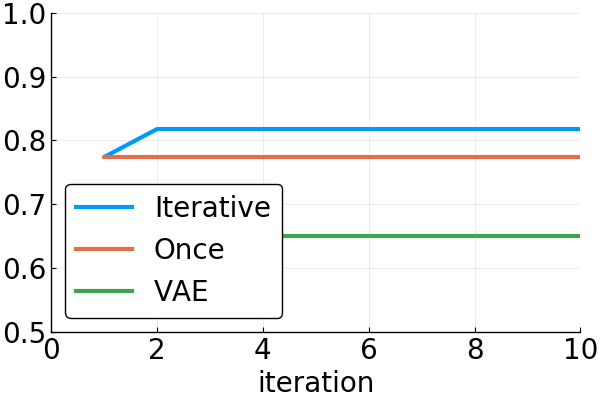} & \includegraphics[width=0.25\textwidth]{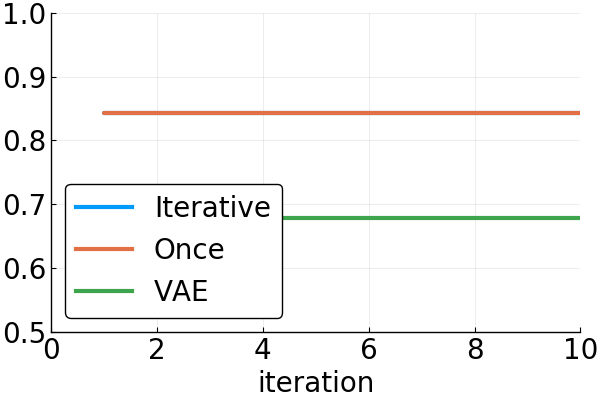} & \includegraphics[width=0.25\textwidth]{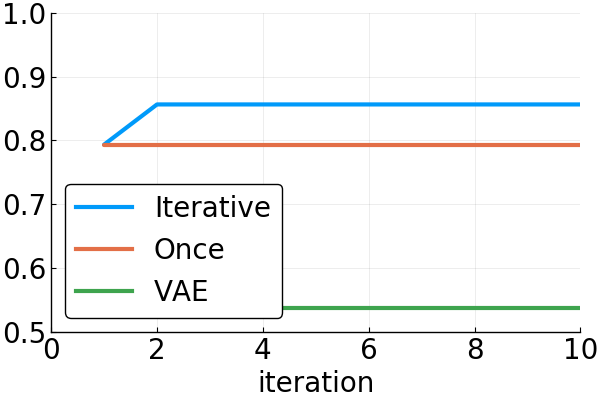} & \includegraphics[width=0.25\textwidth]{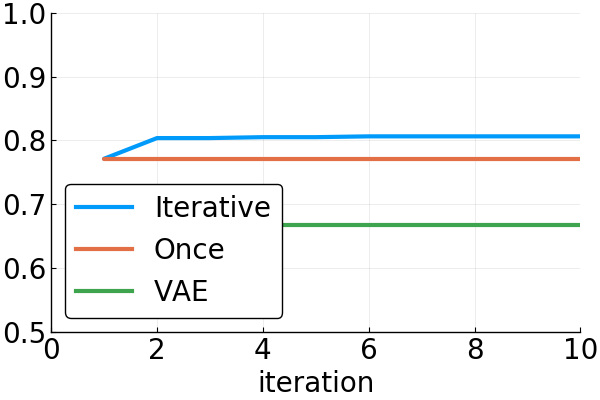}\tabularnewline
 &  & \medskip{}
 & \tabularnewline
\includegraphics[width=0.25\textwidth]{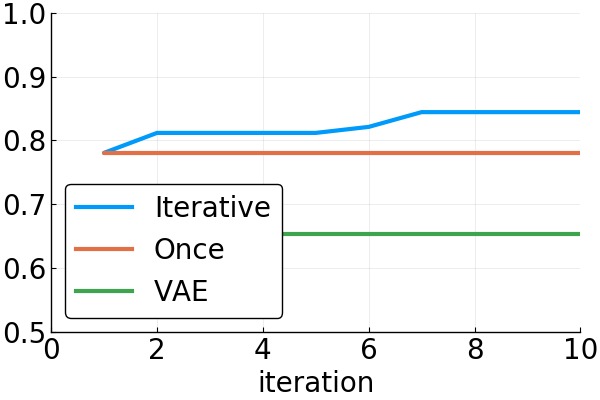} & \includegraphics[width=0.25\textwidth]{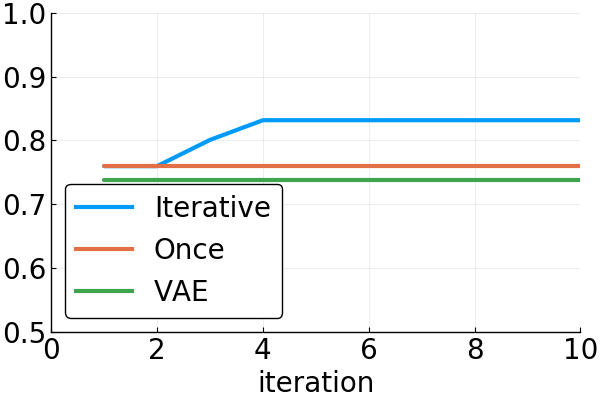} &  & \tabularnewline
 &  & \medskip{}
 & \tabularnewline
\multicolumn{4}{c}{(a) Searching for unseen MNIST digits $\{1,\dots,9,0\}$, from left
to right, top to bottom.}\tabularnewline
 &  & \bigskip{}
 & \tabularnewline
\includegraphics[width=0.25\textwidth]{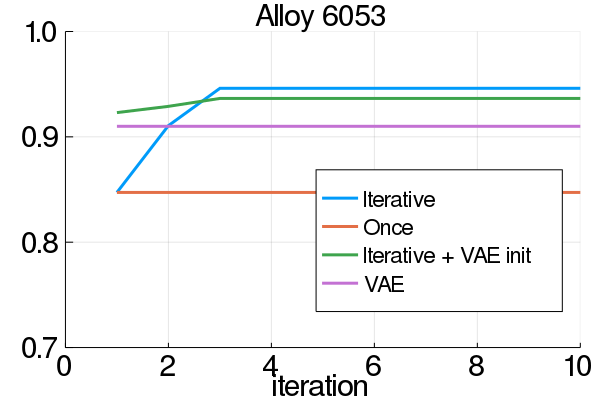} & \includegraphics[width=0.25\textwidth]{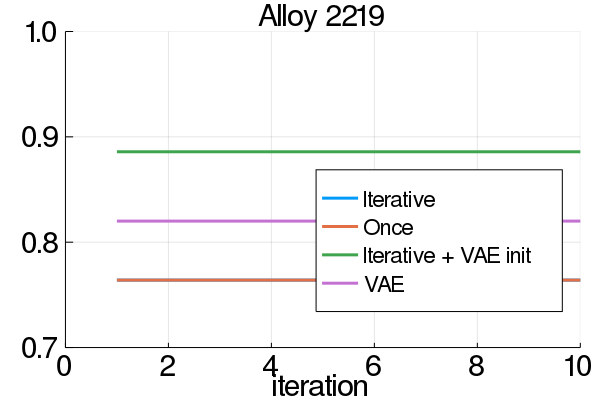} & \includegraphics[width=0.25\textwidth]{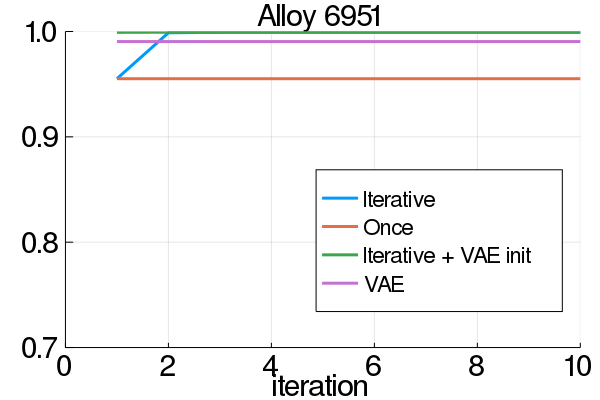} & \includegraphics[width=0.25\textwidth]{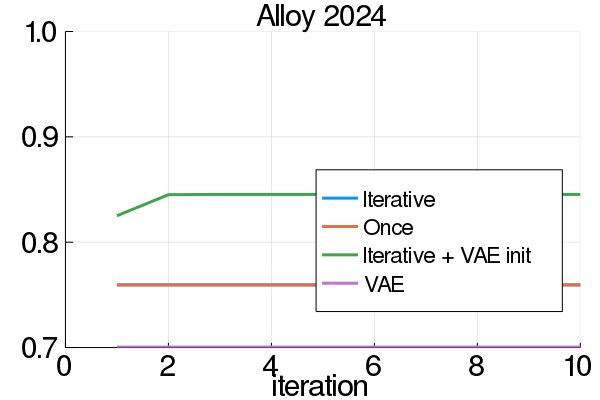}\tabularnewline
 &  & \medskip{}
 & \tabularnewline
\includegraphics[width=0.25\textwidth]{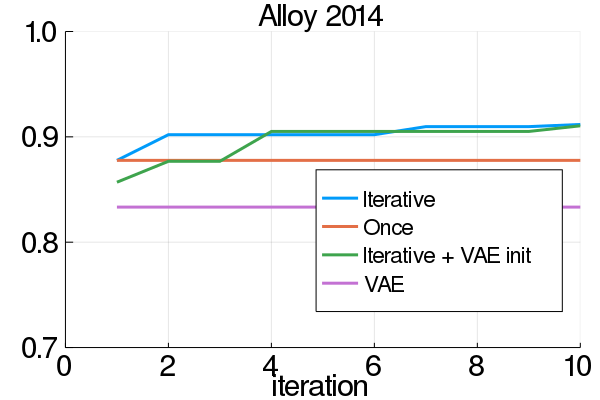} & \includegraphics[width=0.25\textwidth]{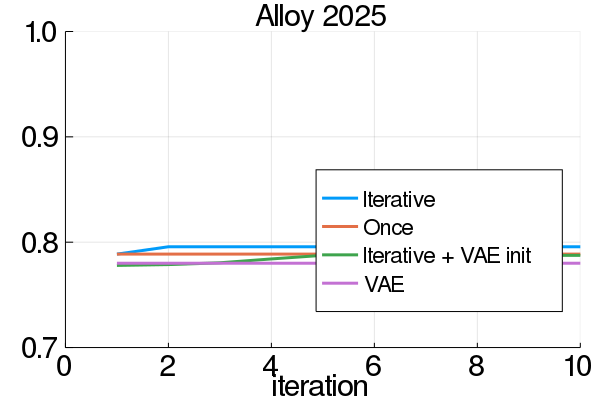} & \includegraphics[width=0.25\textwidth]{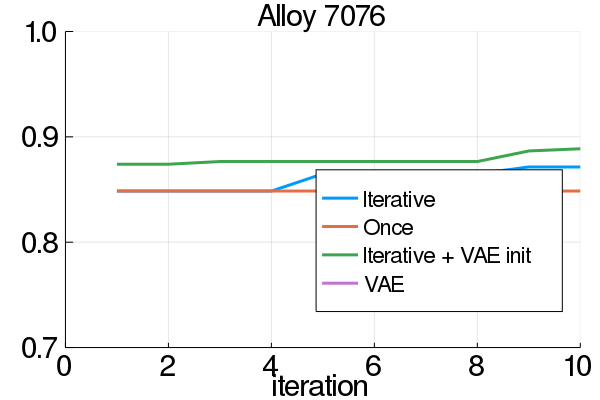} & \includegraphics[width=0.25\textwidth]{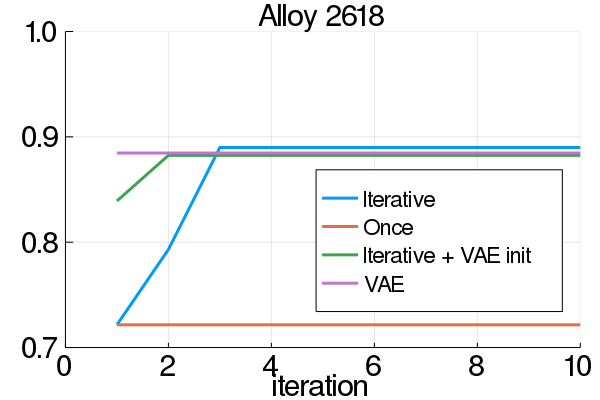}\tabularnewline
 &  & \medskip{}
 & \tabularnewline
\multicolumn{4}{c}{(b) Searching for unseen Alloys.}\tabularnewline
 &  & \medskip{}
 & \tabularnewline
\end{tabular}\endgroup

\caption{Searching for: (a) unseen digits, and (b) unseen alloys designs. Cosine
distance between target and best found vs iterations. Best viewed
in color. \label{fig:Performance-curves-iterative-explore}}
\end{figure}

\subsection{Novelty discovery}

We demonstrate the effectiveness of HyperVAE+BO for finding realistic
designs close to an ideal design, which lies outside known design
classes. The performance measure is how close we get to the given
ideal design, as measured in cosine distance for simplicity.

In each of the following two experiments, the BO objective is to search
for a novel unseen design $x^{*}$, an unseen digit or alloy, by maximizing
a Cosine distance $F(x^{*})$. The maximum number of BO iterations
is set to 300 and the search space is $[-5,5]$ for each $z$ and
$u$ dimension.

\subsubsection{Digit discovery}

This experiment illustrates the capability of HyperVAE+BO in novel
exploration on MNIST. For each experiment, one digit is held out.
We used nine digit classes for training and tested the model ability
to search for high quality digits of the remaining unseen digit class.
BO is applied to search for new digits that are similar to a given
new exemplar in the $z$-space.

In the iterative process, an empty image $d_{1}=\boldsymbol{0}$ is
given at the first step, and subsequently updated as $d_{t}=x_{t-1}^{*}$.
After each step $t$ we set $u_{t}=\mu(x_{t}^{*})$. The quality curves
are presented Fig.~\ref{fig:Performance-curves-iterative-explore}
(a). Examples of discovery process are listed in Fig.~\ref{fig:Best-digits-found}.
The figures show that VAE has a very limited capability to support
exploration outside the known regions, while HyperVAE is much more
flexible, even without the iterative process (\#Step = 1). With more
iterative refinements, the quality of the explored samples improves.

\subsubsection{Alloy discovery}

We now use the framework to search for a new class of alloys. For
each experiment, one alloy is held out. Models are trained on the
remaining 29 alloys. We work on the phase space as a representation
of the material composition space, to take advantage of the closeness
of phase space to the target performance. We treat the phase diagrams
as matrices whose values are proportions of phases at different temperatures.
The goal is to search for a new class of alloys that is similar to
the ``ideal'' alloy that has not been seen in any previous alloy
classes. BO is applied to search for new alloys that are similar to
a given new ideal alloy in the space of $z$. In the iterative process,
we can initialize the search by an uninformative model $u_{1}=\boldsymbol{0}$
or the one found by VAE+BO (the ``Iterative + VAE init''). Subsequently
the model is updated by setting $d_{t}=x_{t-1}^{*}$. The $u$ variable
is set to $u_{t}=\mu(x_{t}^{*})$ after each step $t$.

We utilize the matrix structure of the phase diagram and avoid overfitting
by using matrix representation for the input \cite{do2018learning}.
To inversely map the phase diagram back to the element composition,
we use the inverse program learned from the phase-composition dataset,
as described in \cite{nguyen2019incomplete}. To verify that the found
materials are realistic (to account for the possible error made by
the inverse program), we run the Thermo-Calc software to generate
the phase diagrams. These computed phase diagrams are compared against
the discovered phase diagrams. The result from Thermo-Calc confirms
that the found alloys are in the target class.

To examine the effect of initialization to HyperVAE+BO performance,
we initialized it by either uninformative hypothetical alloy (e.g.,
with hyper prior of zeros), with the alloy found by VAE+BO, or with
a chosen known alloy. The performance curves are shown in Fig.~\ref{fig:Performance-curves-iterative-explore}
(b). ``Once'' means running HyperVAE for just one step. ``Iterative
+ VAE init'' means initialization of $d_{1}=x^{*}$ by VAE. It shows:
(a) For a majority of cases, HyperVAE+BO initialized uninformatively
could find a better solution than VAE+BO, and (b) initializing HyperVAE+BO
with solution found by VAE+BO boosts the performance further, sometimes
a lot more. This suggests that care must be taken for initializing
HyperVAE+BO.

We examine the results of the ten most difficult to find alloy targets,
i.e. the alloys whose distance to their nearest alloy are largest,
in descending order of difficulty. Table~\ref{tab:Found-target}
shows the element composition errors of found alloys. The results
show that most alloys are found to be in the target class and all
found alloys are close to the boundaries of their targets (at $\pm20\%$).
Table~\ref{tab:Found-target} also shows that the Thermo-Calc phase
calculation agrees with the predicted phase, i.e. small errors. The
alloy 6951 and 6463 have the smallest errors compared to others.

\begin{table}
\caption{\textbf{Column A} - Element composition errors of found alloys (the
composition is predicted by the method in \cite{nguyen2019incomplete}).
The found alloys are expected to reside within $\pm20\%$ relative
error to the target alloy to stay within its class. \textbf{Column
B} - Verification of phase in Thermo-Calc simulator, where the phase
error is calculated as the mean error relative to the maximum proportion
of each phase. The errors of the best method are reported. Alloys
are ranked by their relative distance to the nearest neighbor, in
decreasing order. \label{tab:Found-target}}

\centering{}%
\begin{tabular}{|>{\centering}p{0.2\columnwidth}|>{\centering}p{0.3\columnwidth}|>{\centering}p{0.3\columnwidth}|}
\hline
Alloy & A - Element error (\%) & B - Phase error (\%)\tabularnewline
\hline
\hline
6053 & \textbf{11.3} & 3.0\tabularnewline
\hline
2219 & 26.6 & 3.0\tabularnewline
\hline
6951 & \textbf{20.0} & 0.4\tabularnewline
\hline
2024 & 31.5 & 3.5\tabularnewline
\hline
2014 & \textbf{18.5} & 1.4\tabularnewline
\hline
2025 & \textbf{4.4} & 2.9\tabularnewline
\hline
7076 & 31.8 & 2.5\tabularnewline
\hline
2618 & 23.9 & 3.6\tabularnewline
\hline
\end{tabular}
\end{table}

\section{Related Work}

Our method can be considered as a lossless compression strategy where
the HyperVAE compresses a family of networks that parameterize the
parameters of distributions across datasets. The total code length
of both the model and data misfits are minimized using HyperVAE, thus
help it generalize to unseen data. This is in contrast to the lossy
compression strategy \cite{chen2016variational} where local information
of images are freely decoded independent of the compressed information.

The HyperVAE shares some insight with the recent MetaVAE \cite{choi2019meta},
but this is different from ours in the target and modeling, where
the latent $z$ is factored into data latent variable $z$ and the
task latent variable $u$. HyperVAE is related to Bayesian VAE, where
the model is also a random variable generated from some hyper-prior.
There has been some work on priors of VAE \cite{le2018variational,tomczak2017vae},
but using VAE as a prior for VAE is new.

HyperGAN \cite{ratzlaff2019hypergan} is a recent attempt to generate
the parameters of model for classification. This framework generates
all parameters from a single low dimension Gaussian noise vector.
Bayesian neural networks (BNN) in \cite{wang2018adversarial} also
use GAN framework for generating network parameters $\theta$ that
looks real similar to one drawn from BNN trained with stochastic gradient
Langevin dynamics. However, GAN is not very successful for exploration
but more for generating realistic samples.

Continual learning are gaining ground in recent years. Variational
continual learning \cite{nguyen2018variational}, for example, solves
catastrophic forgetting problems in supervised learning, but it still
needs a set of prototype points for old tasks. \cite{rao2019continual}
tackles this problem in unsupervised tasks and also does task inference
as ours, however our settings and approaches are different. Meta-learning
frameworks for classification and regression \cite{finn2018meta,yoon2018bayesian,mishra2018simple}
is another direction where the purpose is to learn agnostic models
that can quickly adapt to a new task.

\section{Conclusion}

We proposed a new method called HyperVAE for encoding a family of neural
network models into a simple distribution of latent representations.
A neural network instance sampled from this family is capable of modeling
the end task in which the family is trained on. Furthermore, by explicitly
training the variational hyper-encoder network over a complex distribution
of tasks, the hyper-network learns the smooth manifold of the family
encoded in the posterior distribution of the family. This enables
the model to extrapolate to new tasks close to trained tasks, and
to transfer common factors of variation across tasks. In the handwritten
digit example, the transferable factors may include writing styles,
font face and size. It can be thought of as expanding the support
of the distribution of trained model, thus is useful for downstream
tasks such as searching for a data distribution close to existing
ones and reducing the false positive error in outlier detection.

\section*{Acknowledgments}
This research was partially funded by the Australian Government through the Australian Research Council (ARC). Prof Venkatesh is the recipient of an ARC Australian Laureate Fellowship (FL170100006).

\bibliographystyle{plain}
\bibliography{ME,phuoc,truyen}

\end{document}